\pdfoutput=1

\documentclass[11pt]{article}

\usepackage[]{ACL2023}

\usepackage{times}
\usepackage{latexsym}

\usepackage[T1]{fontenc}

\usepackage[utf8]{inputenc}

\usepackage{microtype}

\usepackage{inconsolata}

\usepackage{graphicx}
\usepackage{amsmath}
\usepackage{amssymb}
\usepackage{amsfonts}
\usepackage{booktabs}
\usepackage{algorithm}
\usepackage{algorithmic}
\usepackage{bm}
\usepackage{hyperref}
\usepackage{multirow}
\usepackage{makecell}
\usepackage{tablefootnote}

\title{A Survey on Natural Language Processing for Programming}

\author{Qingfu Zhu$^\sharp$, Xianzhen Luo$^{\sharp }$, Fang Liu$^{\flat}$, Cuiyun Gao$^{\natural}$, Wanxiang Che$^{\sharp}$\thanks{\ \ Corresponding author.}\\
  $^\sharp$Harbin Institute of Technology, Harbin, China \\
  $^\flat$Beihang University, Beijing, China \\
  $^\natural$Harbin Institute of Technology, ShenZhen, China \\
  {\tt \{qfzhu,xzluo,car\}@ir.hit.edu.cn} \\
  {\tt fangliu@buaa.edu.cn} \\
  {\tt gaocuiyun@hit.edu.cn} \\
  }

\begin{document}
\maketitle
\begin{abstract}
Natural language processing for programming aims to use NLP techniques to assist programming.
It is increasingly prevalent for its effectiveness in improving productivity.
Distinct from natural language, a programming language is highly structured and functional.
Constructing a structure-based representation and a functionality-oriented algorithm is at the heart of program understanding and generation.
In this paper, we conduct a systematic review covering tasks, datasets, evaluation methods, techniques, and models from the perspective of the structure-based and functionality-oriented property, aiming to understand the role of the two properties in each component.
Based on the analysis, we illustrate
unexplored areas and suggest potential directions for future work.
\end{abstract}

\section{Introduction}
Natural language processing for programming (NLP4P) is an interdisciplinary field of NLP and software engineering (SE), aiming to use NLP techniques for assisting programming \cite{lachmy2021proceedings}.
It could relieve developers from laborious work, e.g., by automatically writing a document for a program.
Meanwhile, it provides easy access for non-professional users to improve efficiency, e.g., by performing a cross-application operation with natural language (NL) interface~\cite{liu2016latent}.
Therefore, it is beneficial for improving the productivity of the whole society.

    \begin{figure}[!t]
        \centering
        \includegraphics[width=215pt]{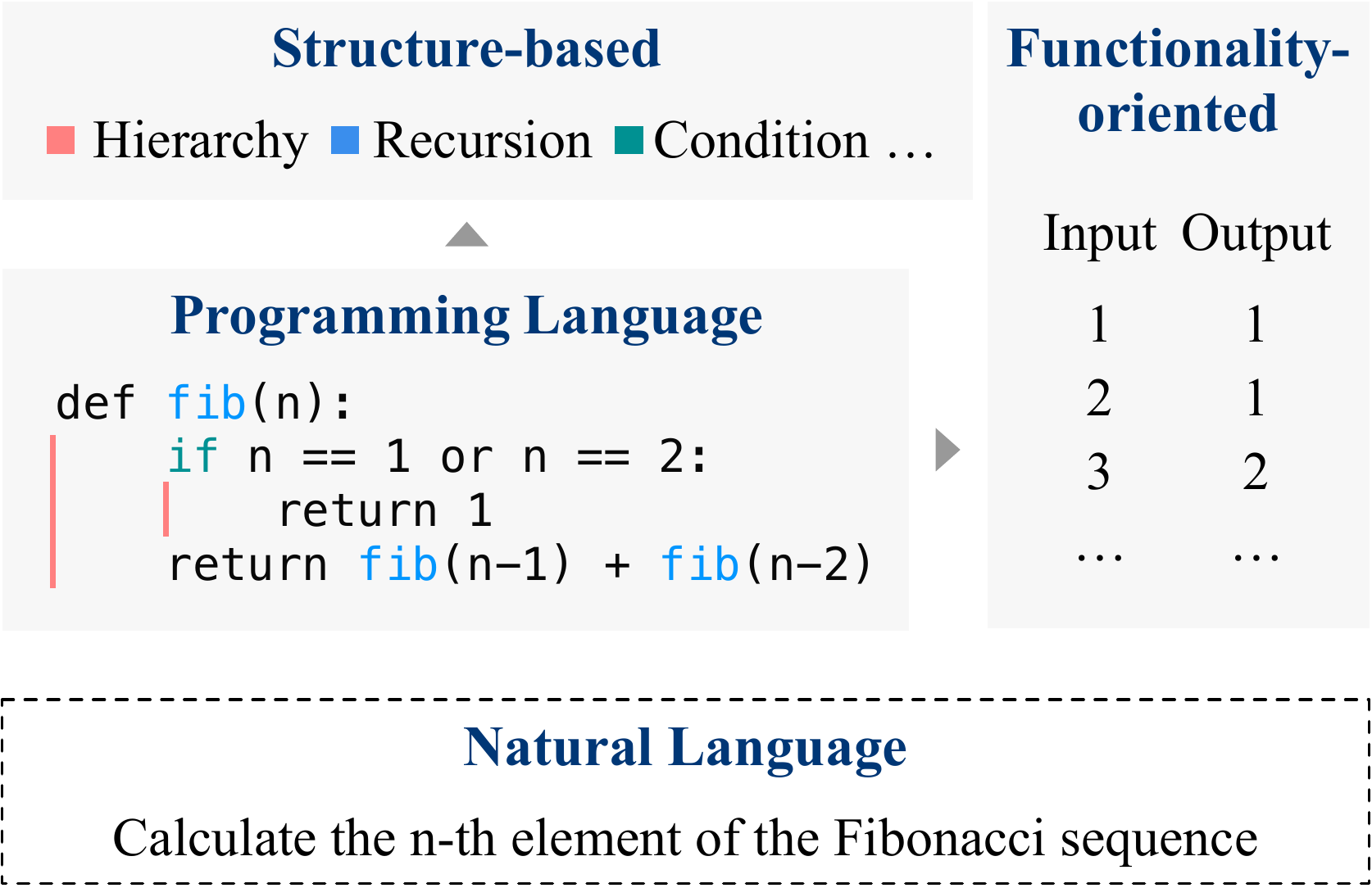}
        \caption{\label{fig:characteristics} 
            An example of the structure-based and functionality-oriented property of programming language.
            Colored fonts and indents denote different aspects of the structure-based property.
        }
    \end{figure}
    
Distinct from NL, a programming language (PL) is characterized by two properties: {\bf structure-based} and {\bf functionality-oriented}, as shown in Figure~\ref{fig:characteristics}.
First, PL is highly structure-based since it intrinsically contains multiple sophisticated structures, such as hierarchy, loops, and recursions.
Appropriately modeling the components and obtaining a structure-based representation is the key to program understanding~\cite{mou2016convolutional,allamanis2018learning,hu2018deep,guo2020graphcodebert,wang2021codet5,guo-etal-2022-unixcoder}.
Second, PL is functionality-oriented since it is executable and ought to convert given input into expected output.
Developing an algorithm oriented to the functionality is at the heart of generating a logically correct program~\cite{chen2021evaluating,hendrycks2021measuring,li2022competition,nijkamp2022conversational,le2022coderl}.
Despite the benefits, the two properties cannot be directly modeled by conventional NLP approaches due to the heterogeneity between NL and PL, making the integration of the properties a fundamental topic in NLP4P.

From the perspective of pre-training,
\citet{niu2022deep} has summarized the recent advance.
Nevertheless, the role of the structure-based and functionality-oriented property has not been sufficiently discussed.
In this paper, we focus on the properties and systematically review their effect in defining tasks (\S\ref{sec:tasks}), constructing datasets (\S\ref{section:data}), forming evaluation methods (\S\ref{section:evaluation}), supporting techniques (\S\ref{section:techniques}), and achieving SOTA performance (\S\ref{section:models}).
Based on the analysis, we further illustrate unexplored areas of current NLP4P and potential directions for future work (\S\ref{section:directions}).
The contributions of this paper are summarized as follows:
\begin{itemize}
\item We identify two properties of PL: structure-based and functionality-oriented, which are essential for program understanding and generation, respectively.
\item From the perspective of the properties, we systematically review current work, covering tasks, datasets, evaluation methods, techniques, and representative models that achieve SOTA performance.
\item By analysis of current NLP4P, we illustrate unexplored areas and suggest potential directions for future work.
\end{itemize}

    \begin{figure}[!t]
        \centering
        \includegraphics[width=215pt]{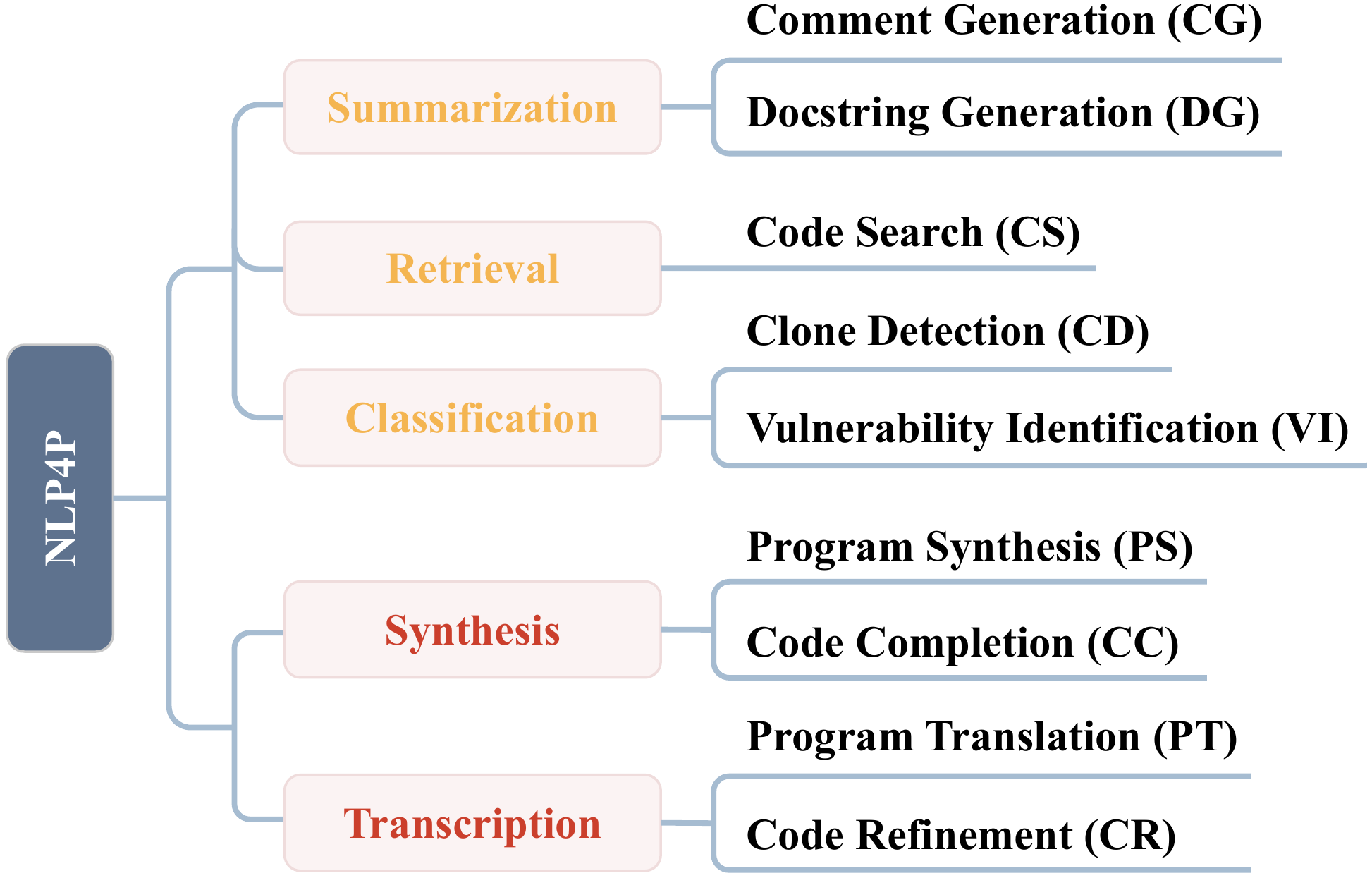}
        \caption{\label{fig:taskoverview} 
            Categories of NLP4P tasks. Orange and red fonts denote the structure-based and functionality-oriented tasks, respectively.
        }
    \end{figure}

\section{Tasks}
\label{sec:tasks}
As shown in Figure \ref{fig:taskoverview}, we classify a task as functionality-oriented if it aims at program generation; otherwise as structure-based.
Within each category, we further divide the tasks according to application scenarios to cluster related tasks and highlight subtle differences between them.
The partition of the structure-based and functionality-oriented roughly aligns with the partition of understanding and generation in NLP.
An exception is the summarization task, which is abstractive and regarded as a generation task in NLP.
We classify it as structure-based since PL lies in its input side, and the key to the task is understanding the content of PL by the structure.

\subsection{Summarization Tasks}
The summarization task summarizes a program into an NL description.
It is crucial for the maintenance of software, especially those involving multiple developers.
According to the format of the output, it can be further divided into \textbf{comment generation} \cite{nie2022impact} and \textbf{docstring generation} \cite{clement2020pymt5}.
The output of the latter contains some structural information, such as parameters and input/output examples.

\subsection{Retrieval Tasks}
The retrieval task mainly refers to the \textbf{code search}.
It aims to retrieve relevant programs given NL query~\cite{husain2019codesearchnet}.
It has a similar application scenario and input/output format to program synthesis. The difference is that its output is extracted from existing programs, rather than being synthesized from scratch.

\subsection{Classification Tasks}
The classification task detects whether given programs have specific characteristics,
e.g., being cloned (\textbf{clone detection}), or being vulnerable (\textbf{vulnerability identification}).
They are essential in protecting software from the effects of ad-hoc reuse~\cite{svajlenko2014towards} and cyber attacks~\cite{zhou2019devign}.
The granularity of the input ranges from a coarse-grained software
repository~\cite{hovsepyan2012software} to a fine-grained function~\cite{russell2018automated,zhou2019devign}.

Despite the fact that NL does not explicitly occur in either input or output, we include tasks of such form for two reasons.
First, PL has been demonstrated to contain abundant statistical properties similar to NL~\cite{mou2016convolutional}.
Second, most of the ways that PL is processed are derived from NLP, like machine translation techniques~\cite{tufano2019empirical} in the transcription task (\S~\ref{sec:transcription}).

\subsection{Synthesis Tasks}
The synthesis task generates a program given a context (which can be NL, PL, or their mixture), thus can accelerate the development process.
It can be further divided into program synthesis and code completion by the formal completeness of the output.
The output of program synthesis is a relatively independent unit, such as a function and a class, while the output of code completion is less restricted, ranging from tokens to code snippets.

\paragraph{Program synthesis} is also called code generation.
It is the systematic derivation of a program from a given specification \cite{manna1980deductive}.
Conventional deductive approaches~\cite{manna1980deductive,polozov2015flashmeta} take logical specifications, which are logically complete but hard to write.
Inductive approaches~\cite{lieberman2001your} list input-output examples as specifications, which are more accessible but incomplete.
In contrast, an NL specification is sufficient to describe the logic of a program.
Meanwhile, it is compatible with input-output examples by including them in a docstring.
Therefore, it can take advantage of both the deductive and inductive approaches.

\paragraph{Code completion} is also called code suggestion in early research~\cite{tu2014localness,hindle2016naturalness}. It suggests the next program token given a context and has been widely applied to IDEs~\cite{li2018code}.
The application scenario includes the completion of method calls, keywords, variables, and arguments.
With the bloom of the pre-trained models, the scenario has been extended to punctuations, statements, and even code snippets~\cite{svyatkovskiy2020intellicode}, further blurring the line between program synthesis and code completion.

\subsection{Transcription Tasks} \label{sec:transcription}
The transcription task converts a given program to meet a specific requirement.
Concretely, \textbf{program translation} aims to convert between high-level PL~\cite{roziere2020unsupervised,zhu2022multilingual}, e.g., C\# and Java.
It can accelerate the update of projects written by deprecated PL, and the migration of algorithms implemented by various PLs.
\textbf{Code refinement} aims to convert a buggy program into correct one~\cite{wang2021codet5}.
It is closely related to vulnerability identification but is required to fix the detected bugs simultaneously.
The transcription task differs from the synthesis task in two aspects.
First, its input program is formally complete (input program is None or a function header in program synthesis, a partial code snippet in code completion).
Second, its output can be strictly aligned with the input in both the format and the content.

\begin{table*}[!ht] \small
\centering
\begin{tabular}{cc|clccrr}
\hline
\multicolumn{2}{c|}{\textbf{}}
& \textbf{Abbr.} &
\textbf{Data} & \textbf{Source} & \textbf{PL} &
\textbf{Size} &
\textbf{Type} \\
\hline
\multicolumn{2}{c|}{\multirow{16}*{\rotatebox{90}{\bf{General Dataset}}}}
& 
\multirow{2}*{pc}& \multirow{2}*{\citealp{christopoulou2022pangu}} & \multirow{2}*{GitHub} & \multirow{2}*{Python} &
\multirow{2}*{147} &
NL-PL\\
\cline{8-8}
& & & & & & & PL\\
\cline{3-8}

& &\multirow{2}*{tp}& \multirow{2}*{THEPILE~\citeyearpar{gao2020pile}} & \multirow{2}*{GitHub, ArXiv,...} & \multirow{2}*{-} & \multirow{2}*{825} & NL\\
\cline{8-8}
& & & & & & & PL\\
\cline{3-8}

& &\multirow{2}*{pb}& \multirow{2}*{\citealp{ahmad-etal-2021-unified}} & {GitHub,} &Java, & 
\multirow{2}*{655}&
NL\\
\cline{8-8}
& & & & Stack Overflow & Python & & PL\\
\cline{3-8}

& &\multirow{2}*{ic}& \multirow{2}*{\citealp{fried2022incoder}} & GitHub, GitLab, &\multirow{2}*{28} & 
\multirow{2}*{216}&
NL\\
\cline{8-8}
& & & & Stack Overflow & & & PL\\
\cline{3-8}

& & ts & thestack & GitHub & 30 & 3,100 & PL\\
\cline{3-8}
& & ac & \citealp{li2022competition} & GitHub & 12 & 715 & PL\\
\cline{3-8}

& & \multirow{2}*{bq}& \multirow{2}*{BigQuery} &
\multirow{2}*{GitHub} &
C/C++, Go, Java, &
\multirow{2}*{340}&
\multirow{2}*{PL}
\\
& & & & & JS, Python & &\\
\cline{3-8}

& &  bp & BIGPYTHON~\citeyearpar{nijkamp2022conversational} & GitHub & Python & 217 & PL \\
\cline{3-8}

& & cp & CodeParrot & GitHub & Python & 180 & PL \\
\cline{3-8}

& & cx & \citealp{chen2021evaluating} & - & Python & 159 & PL\\
\cline{3-8}

& & gp & GCPY~\citeyearpar{le2022coderl} & GitHub & Python & - & PL \\

\hline

\multicolumn{1}{c|}{\multirow{6}*{\rotatebox{90}{\bf{Struc-based}}}}& \multirow{1}*{\rotatebox{0}{\bf{CG}}}&cn& CodeNN~\citeyearpar{iyer-etal-2016-summarizing} & Stack Overflow & C\#, SQL & <1 & NL-PL \\
\cline{2-8}
\multicolumn{1}{c|}{}& \multirow{2}*{\rotatebox{0}{\bf{CS}}}&
\multirow{2}*{csn}& \multirow{2}*{CodeSearchNet~\citeyearpar{husain2019codesearchnet}} & \multirow{2}*{GitHub} & Go, Java, JS, PHP, &
\multirow{2}*{17} &
NL-PL\\
\cline{8-8}
\multicolumn{1}{c|}{}& & & & & Python, Ruby & & PL\\
\cline{2-8}

\multicolumn{1}{c|}{}& \multirow{2}*{\rotatebox{0}{\bf{CD}}}& bc & BigCloneBench~\citeyearpar{svajlenko2014towards} & SeCold & Java & 2 & PL\\
\cline{3-8}
\multicolumn{1}{c|}{}&  & poj & POJ-104~\citeyearpar{mou2016convolutional} & - & C/C++ & <1 & PL \\
\cline{2-8}

\multicolumn{1}{c|}{}& \multirow{1}*{\rotatebox{0}{\bf{VI}}}& dv & Devign~\citeyearpar{zhou2019devign} & QEMU, FFmpeg & C & <1 & PL \\
\hline

\multicolumn{1}{c|}{\multirow{12}*{\rotatebox{90}{\bf{Functionality-oriented}}}}
&
\multirow{8}*{\rotatebox{0}{\bf{PS}}}&cd& CONCODE~\citeyearpar{iyer-etal-2018-mapping} & GitHub & Java & 13 & NL-PL\\
\cline{3-8}

\multicolumn{1}{c|}{}& &cn& CodeNet~\citeyearpar{puri2021codenet} & AIZU, AtCoder & 55 & 8 & NL-PL \\
\cline{3-8}

\multicolumn{1}{c|}{}& & \multirow{2}*{cc} & \multirow{2}*{CodeContests~\citeyearpar{li2022competition}} & CodeNet, Codeforces, & C/C++, Java, & \multirow{2}*{3} & \multirow{2}*{NL-PL}\\
\multicolumn{1}{c|}{}& & & & 
~\citealp{Caballero_Description2Code_Dataset_2016} & Python & &\\
\cline{3-8}

\multicolumn{1}{c|}{}& &ap& APPS~\citeyearpar{hendrycks2021measuring} & \makecell[c]{Codewars, AtCoder,\\ Kattis, Codeforces} & Python & 1 & NL-PL  \\
\cline{3-8}

\multicolumn{1}{c|}{}& & he & HumanEval~\citeyearpar{chen2021evaluating} & Hand-Craft & Python & <1 & NL-PL\\
\cline{3-8}

\multicolumn{1}{c|}{}& & mb & MBPP~\citeyearpar{DBLP:journals/corr/abs-2108-07732} & Hand-Craft & Python & <1 & NL-PL\\
\cline{2-8}

\multicolumn{1}{c|}{}& \multirow{2}*{\rotatebox{0}{\bf{CC}}}& py & PY150~\citeyearpar{raychev2016probabilistic} & GitHub & Python & <1 & PL\\
\cline{3-8}
\multicolumn{1}{c|}{}&  & gjc & Github Java Corpus~\citeyearpar{allamanis2013mining} & GitHub & Java & <1 & PL\\
\cline{2-8}

\multicolumn{1}{c|}{}& \multirow{1}*{\rotatebox{0}{\bf{PT}}}& ct & CodeTrans & Lucene, POI, JGit, Antlr & Java, C\# & <1 & PL\\
\cline{2-8}

\multicolumn{1}{c|}{}& \multirow{1}*{\rotatebox{0}{\bf{CR}}}& bf & Bugs2Fix~\citeyearpar{tufano2019empirical} & GitHub & Java & 15 & PL\\

\hline

\end{tabular}
\caption{\label{data}
An overview of datasets.
Struc-based denotes the structure-based.
Abbreviations (Abbr.) in upper case and lower case denote tasks (Figure~\ref{fig:taskoverview}) and datasets, respectively.
For datasets that contain numerous kinds of PL, the total number of PLs is reported instead of concrete PL types.
The unit of data size is GB.
NL-PL in the last column denotes a parallel dataset whose all samples contain paired NL and PL.
Detached NL or PL denotes a monolingual dataset whose dominant language is NL or PL.
}
\end{table*}

\section{Datasets} 
\label{section:data}
Datasets are the basis for supporting the learning process of tasks.
We thus classify current datasets into general, structure-based, and functionality-oriented, following the categories of tasks (as shown in Table~\ref{data}).
The general dataset is slightly processed and can be used in the early learning stage of a model regardless of tasks.
The structure-based and functionality-oriented datasets are dedicated data specifically formatted for each task.

\subsection{General vs. Dedicated}
There are two primary sources of general datasets:
1) open-source platforms such as GitHub, GitLab, and SeCold,
2) community-based spaces like Stack Overflow.
The datasets are automatically collected and large in scale, thus can be applied to pre-training to ensure generated PL is grammatically correct and logically valid.
However, sometimes they are noisy and non-informative.
For instance, a commit message like ``update'' is of little substantial content; a code snippet answer might be irrelevant to its question~\cite{iyer-etal-2018-mapping}.

Structure-based datasets are specially formatted to support particular tasks.
Concrete structure information is available via open-source parsers, e.g., Tree-sitter.\footnote{https://tree-sitter.github.io/tree-sitter/}
Most functionality-oriented datasets contain a number of test cases for each sample to verify the functional correctness of synthesized programs.
Therefore, the datasets are typically hand-crafted~\cite{chen2021evaluating,DBLP:journals/corr/abs-2108-07732} or collected from online judge websites~\cite{iyer-etal-2018-mapping,puri2021codenet,hendrycks2021measuring,li2022competition}, including AIZU, AtCoder, Codeforces, Codewars, and Kattis.

\subsection{Parallel vs. Monolingual}
To further explore the potential of datasets outside their original tasks, we divide them into parallel (denoted as NL-PL) and monolingual (denoted as NL or PL).
We define a dataset as parallel if {\bf all} samples include paired NL and PL, otherwise as monolingual.
The type of monolingual datasets is denoted by their dominant language.
For instance, Stack Overflow QA pairs with optional code snippets are denoted as NL, and GitHub programs with optional comments are denoted as PL.
Note that a dataset may consist of multiple subsets of different types, we explicitly list them in the last column of Table~\ref{data}.
Generally, parallel datasets are relatively homogeneous, and thus can support other tasks whose dataset is of the same type.
For example, CodeSearchNet can also be used for comment generation~\cite{lu2021codexglue}.
In contrast, a monolingual dataset is usually task-related with specific labels, making it less transferable to other tasks.

\section{Evaluation Methods}
\label{section:evaluation}
The evaluation metric is also closely related to the task.
Considering that the retrieval and classification tasks are well-defined and their metrics (such as F1, MRR, and accuracy) are universally accepted, we focus on the summarization, synthesis, and transcription tasks, whose evaluation remains an open question.
Concretely, the output of summarization is NL. Thus the evaluation can directly refer to NLP.
For the synthesis and transcription task, whose output is PL, the functionality-oriented property is the main concern, assisted by the structure-based property as an auxiliary.

\subsection{NL Evaluation}
NL evaluation can refer to NLP and be conducted by the following two complementary approaches.

\paragraph{Automatic Evaluation} 
is usually implemented by comparing the n-grams between the predicted output and given references.
Concrete metric includes BLEU~\cite{papineni-etal-2002-bleu}, MENTOR~\cite{banerjee-lavie-2005-meteor}, and ROUGE~\cite{lin-2004-rouge}.
However, limited by the number of references, they might correlate weakly with the real quality ~\cite{liu-etal-2016-evaluate}.
Hence, it is crucial to conduct a human evaluation simultaneously.

\paragraph{Human Evaluation}
consists of several independent dimensions, such as naturalness, diversity, and informativeness.
Common annotation methods include point-wise mode~\cite{iyer-etal-2016-summarizing,shi-etal-2021-cast} and
pair-wise mode~\cite{panthaplackel-etal-2020-learning}.
Human evaluation is more accurate, fine-grained, and comprehensive than automatic evaluation.
However, it is also time-consuming and labor-intensive, and thus can only be conducted on a small subset of the test set.

\subsection{PL Evaluation}
PL evaluation can be conducted by the following two methods using the references and the test cases as evidence, respectively.

\paragraph{Reference based Evaluation} 
Regarding a program as a sequence of tokens,
PL can also be evaluated by n-gram based NL metrics, such as BLEU~\cite{wang2021codet5} and exact match (EM,~\citealp{guo-etal-2022-unixcoder}).
To further capture the structure-based property, \citet{ren2020codebleu} propose the CodeBLEU metric, which takes AST and data flow graph into consideration.
Similar to NL, PL is expressive in that a program can be implemented differently, leading to the same weak correlation issue with a limited number of references.

\paragraph{Test Case based Evaluation}  \citet{hendrycks2021measuring} propose two metrics based on test cases:  Test Case Average and Strict Accuracy.
Suppose there is a single generated program and a varying number of test cases for each sample.
Test Case Average computes the average test case pass rate over all samples.
Strict Accuracy is a relatively rigorous metric.
A program is regarded as accepted if and only if it passes all test cases, and the final Strict Accuracy is the ratio of accepted programs.

Actually, we can generate more than one (e.g., $K$) program for each sample to improve the performance.
In this way, Strict Accuracy regards a sample as accepted if any of the $K$ programs pass all test cases.
Therefore, it is also called $p@k$ in some literature.
The sampling size could be huge, but the number of submissions sometimes is limited, like the competition scenario.
To highlight the difference between the sampling and submission, \citet{li2022competition} further propose the $n@k$ metric, which computes the acceptance ratio when sampling $k$ and submitting $n$ programs.

The test case based evaluation is a remarkable progress, which has already in turn improve the training process via reinforcement learning~\cite{le2022coderl}.
Currently, the associated datasets are only available in program synthesis.
Extending it to other functionality-oriented tasks is expected to gain similar improvement.


    
\section{Techniques}\label{section:techniques}
The heterogeneity between NL and PL requires extra effort in techniques to process programs. 
First, the key to understanding the content of a program is appropriately representing its structure information.
Second, at the heart of program generation is elaborately designing an algorithm to achieve functional correctness.
Therefore, we introduce the techniques by structure-based understanding and functionality-oriented generation, respectively.

\subsection{Structure-based Understanding}
Compared with NL, PL has more sophisticated structures, such as hierarchy, loops, and recursions.
Generally, it would benefit the performance by explicitly representing the structures with appropriate data structure, including relative distance, abstract syntax tree, control flow graph, program dependence graph, and code property graph.

\paragraph{Relative Distance}
typically refers to the distance between two tokens in the source code sequence.
In this way, it can be easily combined into token representations as a feature.
\citet{ahmad-etal-2020-transformer} represent the relative distance as a learnable embedding and introduce it into transformer models by biasing the attention mechanism.
Results show that the relative distance is an effective alternative to AST to capture the structure information.
Based on that, \citet{zugner2021language} further extend the concept of relative distance from textual context to AST.
Jointly training a model with the two types of relative distance achieves further improvement.

\paragraph{Abstract Syntax Tree (AST)} is a tree representation that carries the syntax and structure information of a program~\cite{shi2021cast}.
It simplifies inessential parts (e.g., parentheses) of the parse tree by implying the information in its hierarchy.
Each node of AST has \textbf{arbitrary number of children} organized in a \textbf{specific order}.
Therefore, a lossless representation of AST should capture the two characteristics simultaneously.
Despite that, some AST can be complex with a deep hierarchy~\cite{guo2020graphcodebert}, which delays the parsing time and increases the input length (up to 70\%)~\cite{guo-etal-2022-unixcoder}.


\paragraph{Control Flow Graph (CFG)} represents a program as a graph.
Its node (also called a basic block) contains a sequence of successive statements executed together.
Edges between nodes are directed, denoting the order of execution~\cite{allen1970control}.
CFG makes it convenient to locate specific syntactic structures (such as loops and conditional statements) and redundant statements.

\paragraph{Program Dependence Graph (PDG)} is another graphical representation of a program.
Nodes in PDG are statements and predicate expressions, and edges denote both data dependencies and control dependencies~\cite{ferrante1987program}.
The data dependencies describe the partial order between definitions and usages of variables, and have been demonstrated to be beneficial for program understanding~\cite{krinke2001identifying,allamanis2017smartpaste,allamanis2018learning,guo2020graphcodebert}.
Similar to CFG, control dependencies also model the execution order, but it highlights a statement or a predicate itself by determining edges according to its value~\cite{liu2020retrieval}.

\paragraph{Code Property Graph (CPG)} is a joint graph that merges AST, CFG, and PDG~\cite{yamaguchi2014modeling}.
In this way, it takes advantage of all the representations, and thus can comprehensively represent a program for structure-based tasks, such as vulnerability identification~\cite{zhou2019devign} and comment summarization~\cite{liu2020retrieval}.

In summary, a structure-based representation benefits program understanding.
Among the representations, relative distance takes the most concise form but has the minimum structure information, while CPG is the other extreme.
AST, CFG, and PDG are a balance between conciseness and information capacity.
As a tree representation, AST can be more easily integrated by a backbone model than the graphical CFG and PDG, and thus is the most widely used structure-based representation.

\subsection{Functionality-oriented Generation} \label{section:decoding}
Distinct from NL, PL is executable and ought to convert an input into expected output to implement specific functionality.
To achieve this,
it has developed a sampling-based paradigm, which first samples a large volume of candidates and subsequently selects the desired program by test case based filtering and clustering techniques~\cite{chen2021evaluating,li2022competition,nijkamp2022conversational}.

\paragraph{Sampling}
To ensure good coverage of the desired program,
first, the number of sampling should be as large as possible (up to 1M per problem in \citealp{li2022competition}).
Second, it would be better to employ the standard sampling with temperature or the top-k sampling algorithm, rather than the beam search, whose generated candidates can be pretty similar to each other~\cite{li2016simple}.

\paragraph{Filtering and Clustering}
The resulting programs are subsequently filtered by checking the functional correctness on given test cases~\cite{li2022competition}.
However, the number of programs after filtering can still be huge if there are too many programs sampled.
To fit the scenario where the number of submissions is limited, \citet{li2022competition} propose a clustering strategy.
It first clusters the programs according to their behaviors on generated test cases.
Then it selects and submits a program from the clusters one by one.
This strategy avoids repetitively submitting programs with identical bugs.
Nevertheless, it does not consider the error message feedback after each submission, which might be an interesting direction for future work.

\subsection{Backbone Models}
Most of the functionality-oriented algorithms are model-agnostic and have little impact on the choice of a backbone model.
In this section, we focus on the match between the structure-based representation (e.g., AST) and backbone models.

\paragraph{Recurrent Neural Network}(RNN, \citealp{mikolov2010recurrent}) and its variant LSTM~\cite{hochreiter1997long}
are capable of processing variable-length inputs.
Therefore, it is well-suited for representing NL description~\cite{liu2016latent,weigelt-etal-2020-programming} and PL token sequence~\cite{wei2019code}. 
Meanwhile, it accepts the structure-based representation formatted as a sequence, e.g., the pre-order traversal of AST.
However, such transforms are lossy in that the AST cannot be recovered.
To this end, \citet{hu2018deep} propose a structure-based traversal (SBT) approach, adding parentheses into the sequence to mark hierarchical relationships.
Distinct from SBT that adapts data to a model,
\citet{shido2019automatic} propose a Multi-way Tree-LSTM, which directly takes input as AST.
It first encodes the children of a node with a standard LSTM, and subsequently integrates the results into the node with a Tree-LSTM.


\begin{table*}[!ht] \small
\centering
\begin{tabular}{lccccccccc}
\hline


\multirow{2}*{\textbf{Model}} & \multirow{2}*{\textbf{Size}} &
\textbf{CG/csn}&
\textbf{CS/csn}&
\textbf{CD/bc}&
\textbf{VI/dv}&
\textbf{PS/cd}&
\textbf{CC/py}&
\textbf{PT/ct}&
\textbf{CR/bf}

\\

&
& \textbf{BLEU}
&\textbf{MRR}
&\textbf{F1}
&\textbf{ACC}
& \textbf{BLEU}
& \textbf{ES}
& \textbf{BLEU}
&\textbf{BLEU}

\\
\hline

CodeBERT~\citeyearpar{feng-etal-2020-codebert}
& 125M
& 17.83 & 69.3 & 94.1 & 62.08 & - & - & 79.92 & 91.07 \\

PLBART~\citeyearpar{ahmad-etal-2021-unified}
& 140M
& 18.32 & 68.5 & 93.6 & 63.18 & 36.69 & 68.46 & 83.02 & 88.50\\
\hline

GraphCodeBERT~\citeyearpar{guo2020graphcodebert}
& 125M
& - & 71.3 & 95.0 & - & - & - & 80.58 & \bf{91.31} \\

UniXcoder~\citeyearpar{guo-etal-2022-unixcoder}
& 126M
& 19.30 & \bf{74.4} & \bf{95.2}  & - & 38.23 & \bf{72.00} & - & -\\

CodeT5~\citeyearpar{wang2021codet5}
&220M 
& \bf{19.55} & 71.5  & 95.0 & \bf{65.78} 
& \bf{40.73} & 67.12 & \bf{84.03} & 87.64 \\
\hline

\end{tabular}
\caption{\label{codexglue_result}
The results of structure-based models on CodeXGLUE. Abbreviations in upper case and lower case separated by ``/''denote tasks (Figure~\ref{fig:taskoverview}) and datasets (Table~\ref{data}), respectively. ES denotes Levenshtein edit similarity.
}
\end{table*}

\begin{table}[!ht]\small
\centering
\begin{tabular}{lrrrr}
\hline

\textbf{Model} & \textbf{Size} & \textbf{p@1} & \textbf{p@100}\\
\hline

\multirow{3}*{CodeX~\citeyearpar{chen2021evaluating}} 
& 300M & 13.17 & 36.27\\
& 2.5B & 21.36 & 59.50\\
& 12B & 28.81 & 72.31\\
\hline

\multirow{4}*{CodeGen~\citeyearpar{nijkamp2022conversational}} &
350M & 12.76 & 35.19\\
& 2.7B & 23.70 & 57.01\\
& 6.1B & 26.13 & 65.82\\
& 16.1B & 29.28 & 75.00\\
\hline

\multirow{2}*{\shortstack[l]{AlphaCode~\citeyearpar{li2022competition}}} 
& 302M & 11.60 & 31.80\\
& 1.1B & 17.10 & 45.30\\
\hline

\multirow{2}*{\shortstack[l]{PANGU-\\CODER~\citeyearpar{christopoulou2022pangu}}} 
& 317 M & 17.07 & 34.55 \\
& 2.6B & 23.78 & 51.24 \\
\hline

INCODER~\citeyearpar{fried2022incoder} 
& 6.7B & 15.20 & 47.00 \\
\hline

CodeRL~\citeyearpar{le2022coderl}
&770M & 14.05 & 46.06\\
\hline

\end{tabular}
\caption{\label{generation_model_performance}
The results of functionality-oriented models on HumanEval benchmark.
p@k is the pass rate when sampling $k$ candidate programs. The results of CodeRL are computed using the officially released checkpoint.
}
\end{table}

\paragraph{Convolutional Neural Network} 
(CNN, \citealp{lecun1989backpropagation}) extracts the features by scanning an input with a sliding window and applying stacked convolution and pooling operations on the window.
Both two operations can be parallelized, making CNN more time-efficient than RNN.
CNN in NLP4P usually takes input as execution traces~\cite{gupta2020synthesize}, input-output pairs~\cite{bunel2018leveraging}, and encodes them into an embedding as the output.
Similar to Tree-LSTM, CNN can also be adapted to the structure-based representation.
For instance,
\citet{mou2016convolutional} propose a tree-based convolutional neural network (TBCNN), which encodes AST by a weight-base and positional features.

\paragraph{Transformer}\cite{vaswani2017attention} has a similar interface to RNN. The difference lies in the following two aspects. First, it is more time-efficient by solely depending on the attention mechanism, rather than the recurrent unit.
Second, it can better capture long-term dependencies, which is essential for processing PL since programs can be pretty long~\cite{ahmad-etal-2020-transformer}.

Despite these approaches, some studies explore the usage of feed-forward neural network~\cite{iyer-etal-2016-summarizing,loyola-etal-2017-neural}, recursive neural network~\cite{liang2018automatic}, and graph neural network~\cite{liu2020retrieval}.
The architecture of the models, as well as RNN and CNN, can be flexibly adapted to customized data, e.g., the primitive AST.
While for large-scale general data, the transformer is suggested due to its high capacity and easy access to pre-training.

\begin{table}[!ht]\small
\centering
\begin{tabular}{llcc}
\hline

\textbf{Model} & \textbf{Arch.} & \textbf{Data}  & \textbf{Developer} \\
\hline
CodeBERT & Enc  & csn & Microsoft \\ 
PLBART & Enc-Dec & pb & UCLA \\
\hline
GraphCodeBERT & Enc & csn & Microsoft \\ 
UniXcoder & Enc-Dec & csn & Microsoft \\
CodeT5 & Enc-Dec & csn, bq & Salesforce \\
\hline
CodeX & Dec & cx & OpenAI \\
AlphaCode & Dec & ac, cc, ap & DeepMind \\
PANGU-CODER & Dec & pc & Huawei \\  
INCODER & Dec & ic & Facebook \\  
CodeGen & Enc-Dec &tp, bp, bq & Salesforce \\ 
CodeRL & Enc-Dec & gp, ap & Salesforce \\ \hline

\end{tabular}
\caption{\label{model_intro}
The architecture (Arch.), training dataset, and developer of representative pre-training models.
Enc, Dec, and Enc-Dec denote the encoder-only, decoder-only, and encoder-decoder architecture, respectively.
}
\end{table}

\section{Representative Pre-training Models} \label{section:models}
SOTA pre-training models can be roughly divided into two categories according to the benchmarks they are evaluated.
The first category focuses on the CodeXGLUE benchmark~\cite{lu2021codexglue}, which is composed primarily of structure-based datasets, as shown in Table~\ref{codexglue_result}.
The second aims at passing the test cases of functionality-oriented datasets, as typified by HumanEval~\cite{chen2021evaluating} in Table~\ref{generation_model_performance}.
We denote the two categories as structure-based models and functionality-oriented models.
Table~\ref{model_intro} shows the architecture, training dataset, and developer of the models.



\subsection{Structure-based Models}
The performance of this category is shown in Table~\ref{codexglue_result}.
GraphCodeBERT, UniXcoder, and CodeT5 incorporate data dependencies of PDG, AST, and node types of AST, respectively.
As a reference, we also report the result of an encoder-only based BERT and an encoder-decoder based PLBART, neither of which utilize the structure representation.

Generally, incorporating structure-based representation can boost the performance of NLP4P tasks.
Concretely, UniXcoder performs better on understanding tasks, while CodeT5 outperforms others on generation tasks.
For program synthesis, although CodeT5 has the highest BLEU score, it would be better to use its variant CodeRL or other functionality-oriented models.
In our experiments, the functional correctness measured by p@k of the former is not as good as that of the latter.






\subsection{Functionality-oriented Models}
Table~\ref{generation_model_performance} shows the p@k results of functionality-oriented models on HumanEval.
The performance is primarily supported by the large scale sampling~\cite{chen2021evaluating} and test case based filtering~\cite{li2022competition} in the inference process.
Based on that, CodeRL feedback the execution result of example test cases\footnote{The example test cases are part of the NL description, not the test cases of the test set.} into the fine-tuning process, and it is a model-agnostic approach that can combine with all the models in Table~\ref{generation_model_performance} to improve the performance.
Along this research line, introducing the feedback of given test cases into the upstream pre-training process is expected to gain further improvement.








\section{Future Directions} \label{section:directions}
Taking advantage of both SE and NLP, NLP4P has achieved remarkable performance.
However, some features of the two fields (such as the iterations of SE and multilingual learning in NLP) have not been sufficiently explored.
Incorporating them is expected to further improve the performance.

\subsection{Iterative NLP4P}

Generally, programming is an evolutionary process involving multiple iterations, rather than writing from scratch at one time.
For instance, it is difficult to solve a problem with a single submission despite the developers being experienced.
As a reference, the average accept rate of Codeforce,\footnote{http://codeforces.com/} a competitive programming website, is only 50.03\%.

Nevertheless, most existing models are trained to accomplish their tasks regardless of historical context information.
Taking the program synthesis as an instance, 
once a program fails to satisfy its requirements, it will be re-generated from scratch.
The error message and previous version are not taken into account and efficiently utilized.
Iterative NLP4P, a progressive programming paradigm with a natural language interface, may shed light on this problem.
It has access to the complete historical context and thus can pay more attention to fixing existing problems and avoid introducing new bugs.

\subsection{Multilingual NLP4P}
As the bloom of the open source software platform, e.g., GitHub,
source code, along with their NL descriptions, has accumulated to a considerable amount, making it possible to learn a data-driven NLP4P model.
However, the distribution of these data is highly unbalanced.
Most of the NL part is English, and the PL part is Java and Python.
As a result, the performance of low-resource NL and PL is much worse than the average performance.
For example, Ruby only takes a minor proportion in CodeSearchNet dataset and is inferior to other PL in both code search and comment generation tasks~\cite{feng-etal-2020-codebert}.

To bridge the gap between different languages, the simplest way is to translate a low-resource language into its high-resource counterpart.
For tasks whose input is low-resource NL, we can translate it into English before sending it to the model.
For tasks whose output is low-resource PL, we can first generate a Java program and subsequently translate it into the desired PL.
However, it introduces extra effort and cascading errors during the translation.
Multilingual learning approaches \cite{conneau-etal-2020-unsupervised,liu2020multilingual,xue-etal-2021-mt5} provide access to address the issue.
It can efficiently utilize the data presented in various languages, representing them in a unified semantic space and avoiding cascading errors.

\subsection{Multi-modal NLP4P}
NL specification may refer to other modalities (e.g., figures) for better understanding.
For instance, the ``Seven Bridge Problem'', a classical graph problem, is hard to understand by plain NL descriptions.
At the heart of the multi-modal approaches is the alignment of various modalities.
However, there is no such dataset in the area of NLP4P, and annotating a new one is costly.
Therefore, it would be crucial to utilize the knowledge entailed in the existing multi-modal datasets (e.g., COCO~\citealp{lin2014microsoft}) and the language-vision pre-trained models (such as CLIP~\citealp{radford2021learning}, Flamingo~\citealp{alayrac2022flamingo}, and METALM~\citealp{hao2022language}).

\section{Conclusion}
In this paper, we review a broad spectrum of NLP4P work.
We identify two intrinsic properties of PL: structure-based and functionality-oriented, which are at the heart of program understanding and generation, respectively.
They naturally partition the tasks, datasets, techniques, and models, highlighting the characteristics of each category.
Additionally, the structure-based property is the key to the choice of backbone models,
and the functionality-oriented property is the primary concern of evaluation methods.
Through the analysis, we list topics that have yet to be fully considered and might be worth researching in the future.

\section*{Limitations}
This paper focuses on the intersection of NLP and SE.
Programming approaches whose algorithms are irrelevant to NLP and the input excludes NL are not fully discussed, e.g., the deductive and inductive program synthesis.
Similarly, universal NLP approaches not devoted to programming are less sufficiently introduced, e.g., the mechanism of the encoding and decoding processes.
The recent work whose techniques details are not publicly available yet (such as ChatGPT\footnote{https://openai.com/blog/chatgpt/} and CodeGeeX\footnote{https://github.com/THUDM/CodeGeeX}) is to be extended if more details are released.

\bibliography{anthology,custom}
\bibliographystyle{acl_natbib}

\end{document}